\documentclass[10pt,twocolumn,letterpaper]{article}

\usepackage{iccv}
\usepackage{times}
\usepackage{epsfig}
\usepackage{graphicx}
\usepackage{amsmath}
\usepackage{amssymb}
\usepackage{multirow}
\usepackage{subfigure}

\usepackage[breaklinks=true,bookmarks=false]{hyperref}

\iccvfinalcopy 

\def\iccvPaperID{****} 

\ificcvfinal\fi

\begin{document}

\title{ VarGFaceNet: An Efficient Variable Group Convolutional Neural Network for Lightweight Face Recognition}

\author{Mengjia Yan\\
Horizon Robotics\\
{\tt\small mengjia.yan@horizon.ai}
\and
Mengao Zhao\\
Horizon Robotics\\
{\tt\small mengao.zhao@horizon.ai}
\and
Zining Xu\\
Horizon Robotics\\
{\tt\small zining.xu@horizon.ai}
\and
Qian Zhang\\
Horizon Robotics\\
{\tt\small qian01.zhang@horizon.ai}
\and
Guoli Wang\\
Horizon Robotics\\
{\tt\small guoli.wang@horizon.ai}
\and
Zhizhong Su\\
Horizon Robotics\\
{\tt\small zhizhong.su@horizon.ai}
}

\maketitle
\ificcvfinal\thispagestyle{empty}\fi

\begin{abstract}
	
	To improve the discriminative and generalization ability of lightweight network for face recognition, we propose an efficient variable group convolutional network called VarGFaceNet. Variable group convolution is introduced by VarGNet to solve the conflict between small computational cost and the unbalance of computational intensity inside a block. We employ variable group convolution to design our network which can support large scale face identification while reduce computational cost and parameters. Specifically, we use a head setting to reserve essential information at the start of the network and propose a particular embedding setting to reduce parameters of fully-connected layer for embedding.  To enhance interpretation ability, we employ an equivalence of angular distillation loss to guide our lightweight network and we apply recursive knowledge distillation to relieve the discrepancy between the teacher model and the student model.  The champion of deepglint-light track of LFR (2019) challenge demonstrates the effectiveness of our model and approach. Implementation of VarGFaceNet will be released at https://github.com/zma-c-137/VarGFaceNet soon.
	
\end{abstract}

\section{Introduction}

With the surge of computational resources, face recognition using deep representation has been widely applied to many fields such as surveillance, marketing and biometrics\cite{arcface,sphereface}. However, it is still a challenging task to implement face recognition on limited computational cost system such as mobile and embedded systems because of the large scale identities needed to be classified.

Many work propose lightweight networks for common computer vision tasks such as \textit{SqueezeNet}\cite{iandola2016squeezenet}, \textit{MobileNet} \cite{mobilenet}, \textit{MobileNetV2} \cite{mobilenetv2}, \textit{ShuffleNet} \cite{shufflenet}. \textit{SqueezeNet}\cite{iandola2016squeezenet} extensively uses $1\times1$ convolution, achieving $50\times$ fewer parameters than \textit{AlexNet}\cite{alexnet} while maintains AlexNet-level accuracy on ImageNet. \textit{MobileNet}\cite{mobilenet} utilizes depthwise separable convolution to achieve a trade off between latency and accuracy. Based on this work, \textit{MobileNetV2}\cite{mobilenetv2}  proposes an inverted bottleneck structure to enhance discriminative ability of network. \textit{ShuffleNet}\cite{shufflenet} and  \textit{ShuffleNetV2}\cite{shufflenetv2} uses pointwise group convolution and channel shuffle operations to further reduce computation cost. Even though they cost small computation during inference and achieve good performance on various applications, optimization problems on embedded system still remain on embedded hardware and corresponding compilers \cite{vargnet}. To handle this conflict, \textit{VarGNet} \cite{vargnet} proposes a variable group convolution which can efficiently solve the unbalance of computational intensity inside a block. Meanwhile, we explore that variable group convolution has larger capacity than depthwise convolution with the same kernel size, which helps network to extract more essential information. However, \textit{VarGNet} is designed for general tasks such as image classificaiton and object detection. It decreases spatial area to the half in the head setting to save memory and computational cost, while this setting is not suitable for face recognition task since detailed information of face is necessary. And there is only an average pooling layer between last conv and fully connected layer of the embedding, which may not extract enough discriminative information.

\begin{figure*}
	\centering
	\subfigure[Normal block]{\label{fig:normal_block}\includegraphics[width=0.5\linewidth]{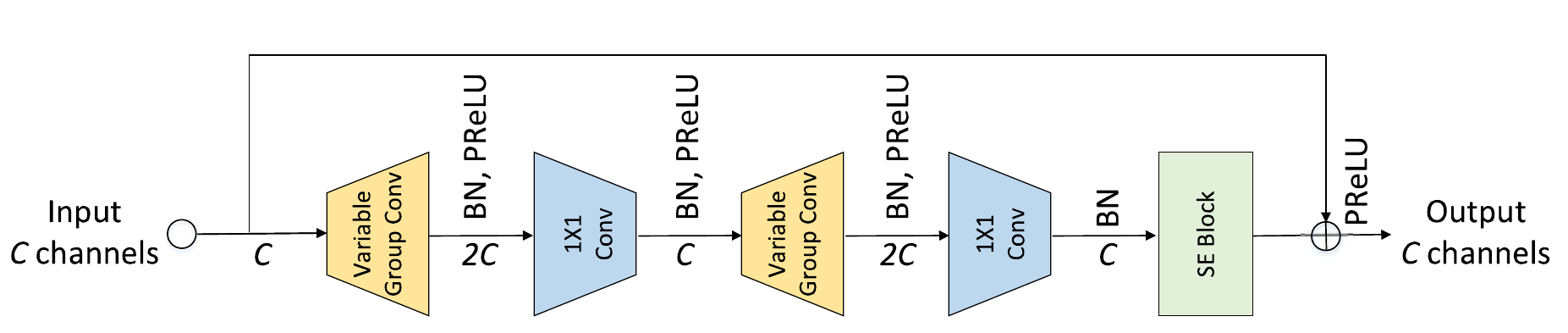}}
	\subfigure[Down sampling block]{\label{fig:d_block}\includegraphics[width=0.48\linewidth]{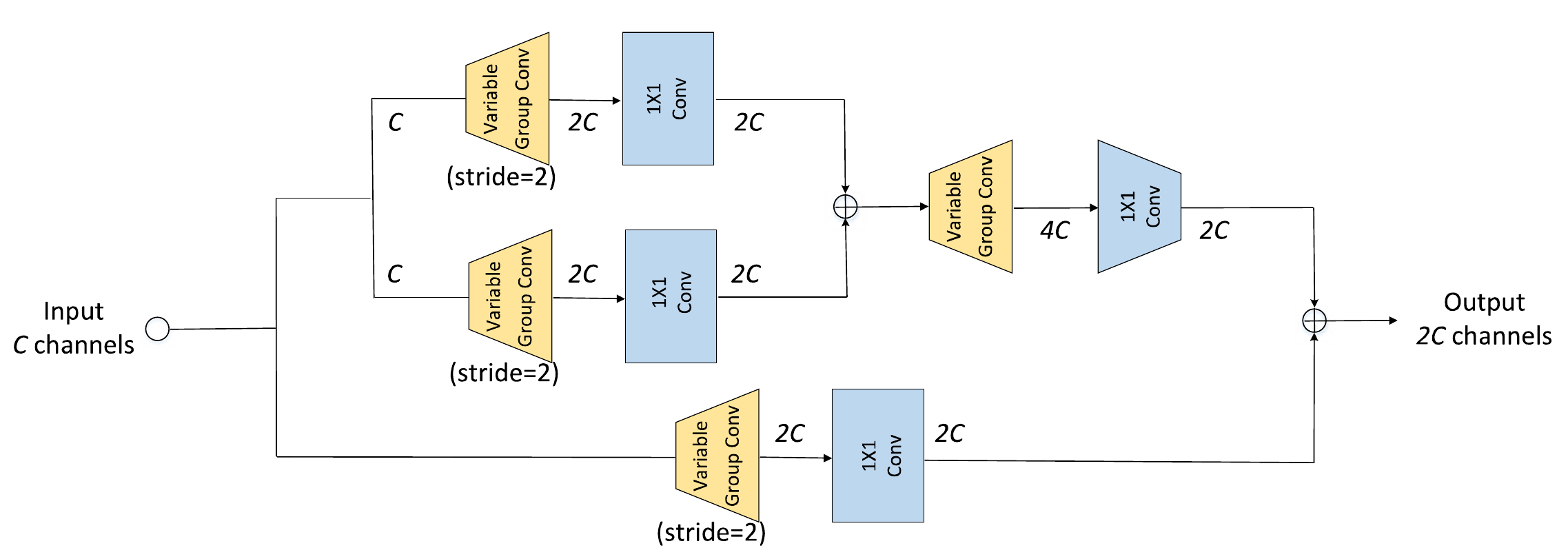}}
	\subfigure[Head setting]{\label{fig:head}\includegraphics[width=0.5\linewidth]{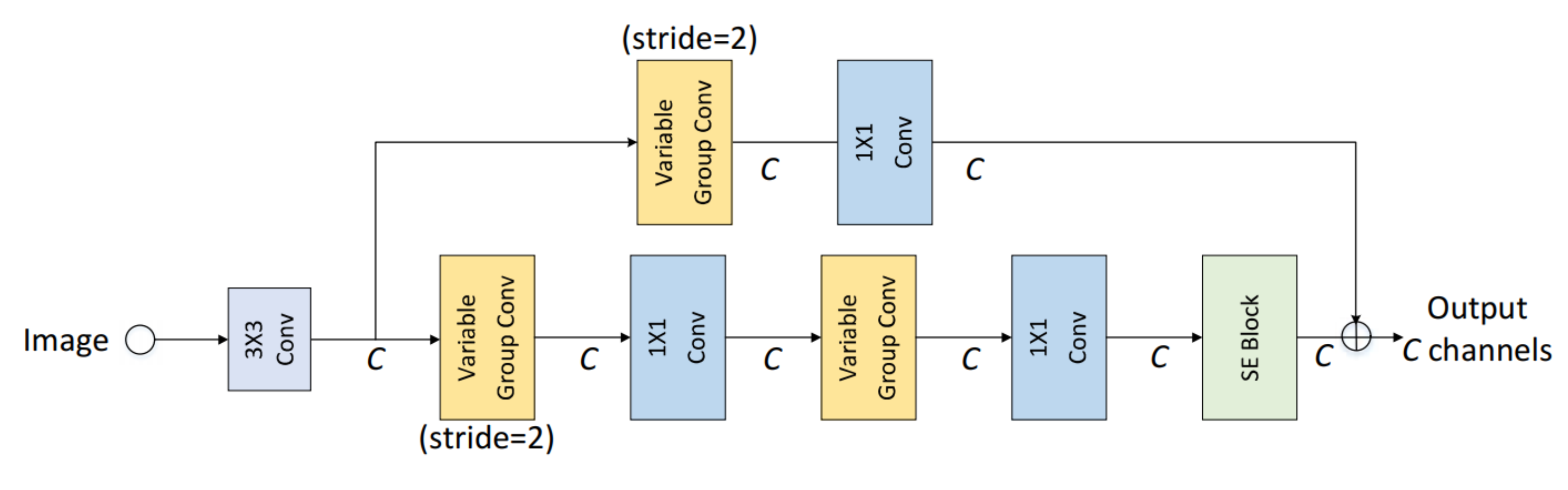}}
	\subfigure[Embedding setting ]{\label{fig:emb} \includegraphics[width=0.45\linewidth]{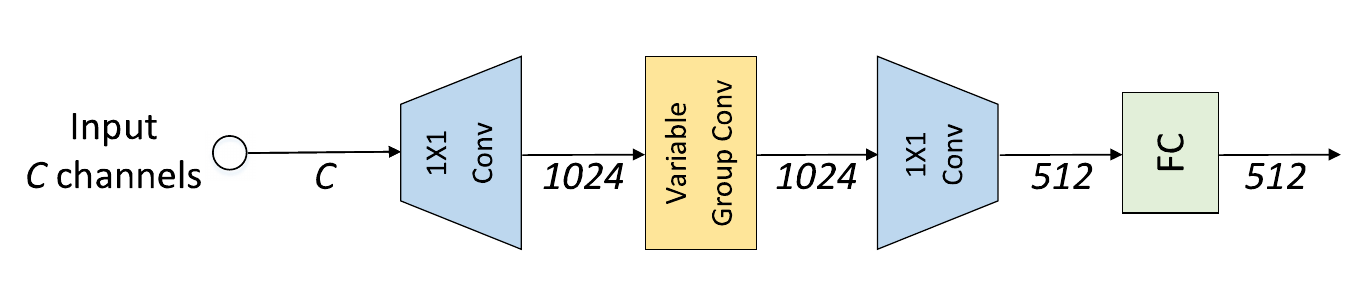}}\\
	\caption{Settings of VarGFaceNet. a) is the normal block of VarGFaceNet. We add SE block on normal block of VarGNet. b) is the down sampling block. c) is head setting of VarGFaceNet. We do not use downsample in first conv in order to keep enough information. c) is the embedding setting of VarGFaceNet. We first expand channels from 320 to 1024. Then we employ variable group convolution and pointwise convolution to reduce the parameters and computational cost while remain essential information. }
	\label{ROC}
\end{figure*}

Based on \textit{VarGNet}, we propose an efficient variable group convolutional network for lightweight face recognition, shorted as VarGFaceNet. In order to enhance the discriminative ability of \textit{VarGNet} for large scale face recognition task, we first add SE block \cite{se} and PReLU \cite{prelu} on blocks of \textit{VarGNet}. Then we remove the downsample process at the start of network to preserve more information. To decrease parameters of network, we apply variable group convolution to shrink the feature tensor to $1\times 1\times 512$ before fc layer. The performance of  VarGFaceNet demonstrates that this embedding setting can preserve discriminative ability while reduce parameters of the network.

To enhance the interpretation ability of lightweight network, we apply knowledge distillation during the training. There are several approaches aim at making the deep network smaller and cost-efficient, such as model pruning, model quantization and knowledge distillation. Among them, knowledge distillation is being actively investigated due to its architectural flexibility. Hinton\cite{hinton2015distilling} introduces the concept of knowledge distillation and proposes to use the softmax output of teacher network to achieve knowledge distillation. To take better advantage of information from teacher network, FitNets\cite{romero2014fitnets} adopts the idea of feature distillation and encourages student network to mimic the hidden feature values of teacher network. After FitNets, there are variant methods attempt to exploit the knowledge of teacher network, such as transferring the feature activation map\cite{heo2018knowledge}, activation-based and gradient-based Attention Maps\cite{yim2017gift}. Recently \textit{ShrinkTeaNet} \cite{duong2019shrinkteanet} introduces an angular distillation loss to focus on angular information of teacher model.  Inspired by angular distillation loss  we employ an equivalent loss with better implementation efficiency as the guide of VarGFaceNet. Moreover, to relieve the complexity of optimization caused by the discrepancy between teacher model and student model, we introduce recursive knowledge distillation which treats the model of student trained in one generation as pretrained model for the next generation.

We evaluate our model and approach on LFR challenge \cite{LFR}. LFR challenge is a lightweight face recognition challenge which requires networks whose FLOPs is under 1G and memory footprint is under 20M. VarGFaceNet achieves the state-of-the-art performance on this challenge which is shown in Section \ref{Experiments}. Our contributions are summarized as follows:

\begin{itemize}
	\item To improve the discriminative ability of \textit{VarGNet} \cite{vargnet} in large-scale face recognition we employ a different head setting and propose a new embedding block. In embedding block, we first expand channels to 1024 by $1\times1$ convolution layer to reserve essential information, then we use variable group conv and pointwise conv to shrink the spatial area to $1\times1$ while saving computational cost. These settings improve the performance on face recognition tasks which shown in Section \ref{Experiments}.
	\item To imporve the generalization ability of lightweight models, we propose recursive knowledge distillation which relieves the generalization gap between teacher models and student models in one generation.
	\item We analyse the efficiency of variable group convolution and employ an equivalence of angular distillation loss during training. Experiments conducted to show the effectiveness of our approach.
\end{itemize}

\section{Approach}

\subsection{Variable Group Convolution}
Group Convolution was first introduced in \textit{AlexNet} \cite{alexnet} for computational cost reduction on GPUs. Then, the cardinality of group convolution demonstrated a better performance than the dimensions of depth and width in \textit{ResNext}   \cite{xie2017aggregated}. Designed for mobile device, \textit{MobileNet} \cite{mobilenet} and \textit{MobileNetV2} \cite{mobilenetv2} proposed depthwise separable convolution inspired by group convolution to save computational cost while keep discriminative ability of convolution. However, depthwise separable convolution spends 95\% computation time in Conv $1\times1$, which causes a large MAdds gap between two consecutive laysers (Conv $1\times1$ and Conv DW $3\times3$) \cite{mobilenet}. This gap is unfriendly to embedded systems who load all weights of the network to perform convolution\cite{xing2019dnnvm}: embedded systems need extra buffers for Conv $1\times1$.

To keep the balance of computational intensity inside a block, \textit{VarGNet} \cite{vargnet} sets the channel numbers in a group as a constant $S$. The constant channel numbers in a group lead to the variable number of groups $n_i$ in a convolution, named variable group convolution. The computational cost of a variable group convolution is:

\begin{equation}
k^2\times h_i \times w_i \times S \times c_{i+1}
\end{equation}

\begin{equation}
S = \frac{c_i}{n_i}
\end{equation}

The input of this layer is $h_i \times w_i \times c_i$ and the output of that is  $h_i \times w_i \times c_{i+1}$. $k$ is the kernel size. When variable group convolution is used to replace depthwise convolution in \textit{MobileNet} \cite{mobilenet}, the computational cost of pointwise convolution is:
\begin{equation}
1^2\times h_i \times w_i \times c_{i+1} \times c_{i+2}
\end{equation}
The ratio of computational cost between variable group convolution and pointwise convolution is $\frac{k^2 S}{c_{i+2}}$ while that between depthwise convolution and pointwise convolution is $\frac{k^2}{c_{i+2}}$. In practice, $c_{i+2} \gg  k^2$, $S>1$, so  $\frac{k^2 S}{c_{i+2}} > \frac{k^2}{c_{i+2}}$. Hence, it will be more computational balanced inside a block when employs variable group convolution  on the bottom of pointwise convolution instead of depthwise convolution.

Moreover, $S>1$ means variable group convolution has higher MAdds and larger network capacity than depthwise convoluiton (with the same kernel size), which is capable of extracting more information.
\subsection{Blocks of Variable Group Network}

Communication between off-chip memory and on-chip memory only happens on the start and the end of block computing when a block is grouped and computed together on  embedded systems \cite{xing2019dnnvm}. To limit the communication cost, \textit{VarGNet} sets the number of output channels to be same as the number of input channels in the normal block. Meanwhile, \textit{VarGNet} expands the $C$ channels at the start of the block to $2C$ channels  using variable group convolution to keep the generalization ability of the block. The normal block we used is shown in Fig. \ref{fig:normal_block}, and down sampling block is shown in Fig. \ref{fig:d_block}. Different from the blocks in \textit{VarGNet} \cite{vargnet}, we add SE block in normal block and employ PReLU instead of ReLU to increase the discriminative ability of the block.

\begin{table*}[htb]
	\begin{center}
		\setlength{\tabcolsep}{7mm}
		\begin{tabular}{l|c|c|c|c|c}			
			\hline
			\hline
			Layer & Output Size & KSize & Stride & Repeat & Output Channels \\
			\hline
			Image & 112x112 & {} & {} & {} & 3 \\
			\hline
			Conv 1 & 112x112 & 3x3 & 1 & 1 & 40 \\
			\hline
			Head Block & 56x56 & {} & 2 & 1 & 40 \\
			\hline
			\multirow{2}{*}{Stage2} & 28x28 & {} & 2 & 1 & \multirow{2}{*}{80} \\
			\cline{2-5}
			& 28x28 & {} & 1 & 2 & {} \\
			\hline
			\multirow{2}{*}{Stage3} & 14x14 & {} & 2 & 1 & \multirow{2}{*}{160} \\
			\cline{2-5}
			& 14x14 & {} & 1 & 6 & {} \\			
			\hline
			\multirow{2}{*}{Stage4} & 7x7 & {} & 2 & 1 & \multirow{2}{*}{320} \\
			\cline{2-5}
			& 7x7 & {} & 1 & 3 & {} \\
			\hline
			Conv 5 & 7x7 & 1x1 & 1 & 1 & 1024 \\
			\hline
			Group Conv & 1x1 & 7x7 & 1 & 1 & 1024 \\
			\hline
			Pointwise Conv & 1x1 & 1x1 & 1 & 1 & 512 \\
			\hline
			FC & {} & {} & {} & {} & 512 \\
			\hline
			\hline
		\end{tabular}
	\end{center}
	\caption{Overall architecture of VarGFaceNet. It only has 1G FLOPs and 5M parameters (memory footprint is 20M saved as float32).}
	\label{tab:vargfacenet}
\end{table*}

\subsection{Lightweight Network for Face Recognition}
\subsubsection{Head setting}
The main challenge of face recognition is the large scale identities involved in testing/training phase. It requires discriminative ability as much as possible to support distinguishing millions of identities. In order to reserve this ability in lightweight networks, we use $3\times3$ Conv with stride 1 at the start of network instead of $3\times3$ Conv with stride 2 in \textit{VarGNet}. It is similar to the input setting of \cite{arcface}. The output feature size of first conv in  \textit{VarGNet} will be downsampled while ours remains the same as input size, shown in Fig. \ref{fig:head}.

\subsubsection{Embedding setting}
To obtain the embedding of faces, many work \cite{arcface,sphereface}  employ a fully-connected layer directly on the top of last convolution. However, the parameters of this fully-connected layer will be huge when output features from last convoluiton are relatively large. For instance, in ResNet 100 \cite{arcface} the output of last conv is $7 \times 7  \times 512$, and the parameters of fc layer (embedding size is 512) are $7 \times 7 \times 512 \times 512$. The overall parameters of fc layer for embedding are  12.25M, and the memory footprint is 49M (float32)!

In order to design a lightweight network (memory footprint is less than 20M, FLOPs is less than 1G), we employ variable group convolution after last conv to shrink the feature maps to $1\times 1 \times 512$ before fc layer. Consequently, the memory footprint of fc layer for embedding is only 1M.  Fig.\ref{fig:emb} shows the setting of embedding block. Shrinking the feature tensor to $1\times1\times512$ before fc layer for embedding is risky since information contains by this feature tensor is limited. To avoid the derease of essential information, we expand channels after last conv to retain as much information as possible. Then we employ variable group convolution and pointwise convolution to decrease the parameters and computational cost while keep information.

Specifically, we first use a $1\times1$ Conv to expand the channels from 320 to 1024. Then we employ a $7\times7$ variable group convolution layer (8 channels in a group) to shrink the feature tensors from $7\times7\times1024$ to $1\times1\times1024$. Finally, pointwise convolution is used to connect the channels and output the feature tensors to $1\times1\times 512$. The new embedding block setting  only takes up 5.78M while the original fc layer takes up 30M ($7\times7\times320\times512$) on the disk.

Experiments of comparison between our network and \textit{VarGNet} in Section \ref{VgR} demonstrate the efficiency of our network on face recognition tasks.

\subsubsection{Overall architecture}
The overall architecture of our lightweight network (VarGFaceNet) is illustrated in Table \ref{tab:vargfacenet}. The memory footprint of our VarGFaceNet is 20M and FLOPs is 1G. We set $S=8$ in a group empirically.  Benefit from variable group convolution, head settings and particular embedding settings, VarGFaceNet can achieve good performance on face recognition task with limited computational cost and parameters. In Section \ref{Experiments}, we will demonstrate the effectiveness of our network on a million distractors face recognition task.

\subsection{Angular Distillation Loss}
Knowledge distillation has been widely used in lightweight network training since it can transfer the interpretation ability of a big network to a smaller network \cite{mobilenet}. Majority tasks that used knowledge distillation are close set tasks \cite{romero2014fitnets, hinton2015distilling}. They apply scores/logits or embeddings/feature magnitude to compute $l2$ distance or cross entropy as loss. However, for open set tasks, scores/logits of training set contain limited information of testing  set and the exact match of featuers maybe over-regularized in some situations. To extract useful information and avoid  over-regularization, \cite{duong2019shrinkteanet} proposes an angular distillation loss for knowledge distillation:
\begin{equation}
L_a(F_t^i,F_s^i) =\frac{1}{N}\sum^{N}_{i=1} || 1- \frac{F_t^i}{||F_t^i||} * \frac{F_s^i}{||F_s^i||} ||^2_2
\label{eq4}
\end{equation}

$F_t^i$ is the $ith$ feature of teacher model, $F_s^i$ is $ith$ features of student model. $m$ is the number of samples in a batch. Eq. \ref{eq4} first computes cosine similarity between features of teacher and student, then minimizes the $l2$ distance between this similarity and 1. Inspired by \cite{duong2019shrinkteanet}, we propose to use Eq. \ref{eq5} to enhance the implementation efficiency. Since cosine similarity is less than 1, minimize Eq. \ref{eq4} is equivalent to minimize Eq. \ref{eq5}.

\begin{equation}
L_s(F_t^i,F_s^i) = \frac{1}{N}\sum^{N}_{i=1} || \frac{F_t^i}{||F_t^i||} - \frac{F_s^i}{||F_s^i||} ||^2_2
\label{eq5}
\end{equation}

Compared with previous $l2$ loss of exact features, Eq. \ref{eq4} and Eq. \ref{eq5} focus on angular information and the distribution of embeddings.

In addition, we employ arcface \cite{arcface} as our classification loss which also pays attention to angular information:

\begin{equation}
L_{Arc}=-\frac{1}{N}\sum^{N}_{i=1}\log\frac{e^{s(cos(\Theta_{y_i}+m))}}{e^{s(cos(\Theta_{y_i}+m))}+\sum^n_{j=1,j\neq y_i}e^{s cos\Theta_j}}
\label{arc}
\end{equation}

To sum up, the objective function we used in training is:

\begin{equation}
L= L_{Arc} + \alpha L_s
\label{loss}
\end{equation}

We empirically set $\alpha = 7$ in our implementation.
\begin{figure*}[htb]
	\begin{center}
		\includegraphics[width=1\linewidth]{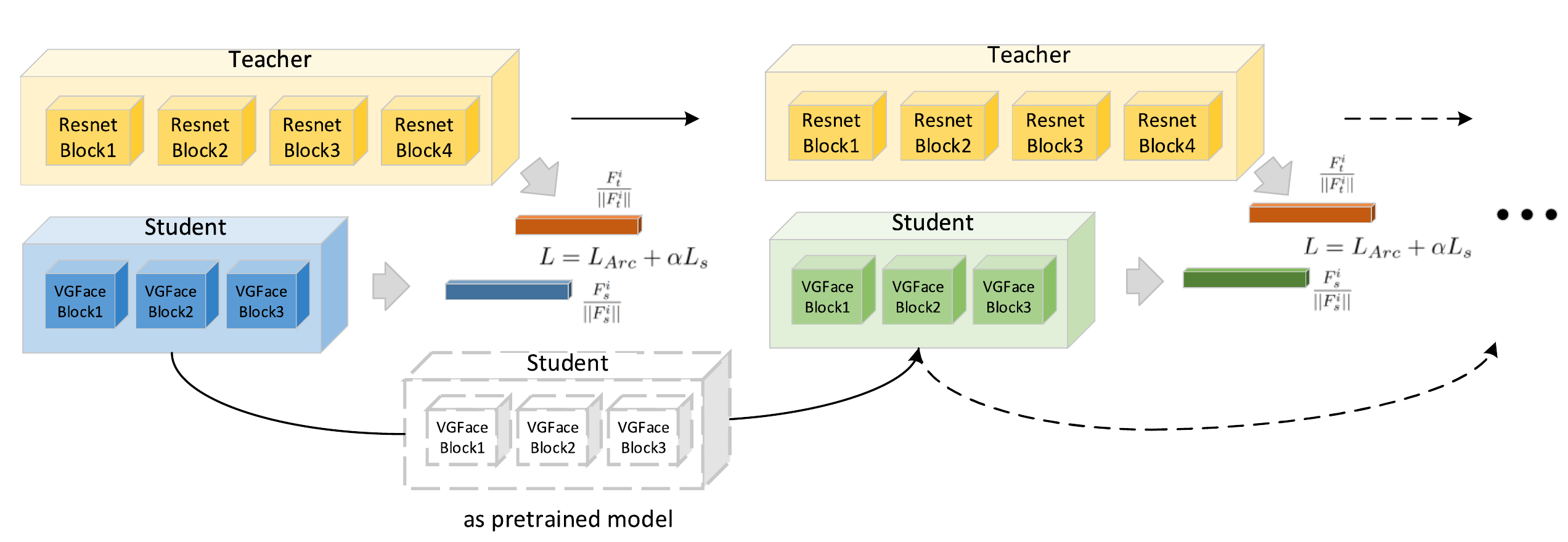}
	\end{center}
	\caption{The process of recursive knowledge distillation. We apply the first generation of student to initialize the second generation of student while the teacher model is remained. Angular distillation loss and arcface loss are used to guide training. }
	\label{kd_f}
\end{figure*}

\begin{table*}
	\begin{center}
		\setlength{\tabcolsep}{5mm}
		\begin{tabular}{l|c|c|c|c|c}
			\hline
			\hline
			Network & LFW & CFP-FP & AgeDB-30 & deepglint-light (TPR@FPR=1e-8) & Flops \\
			\hline
			y2      & 0.99700 & 0.97829 & 0.97517 & 0.803     & 933M \\
			\hline
			VarGFaceNet & 0.99683 & 0.98086 & 0.98100 & 0.855     & 1022M  \\
			\hline
			\hline
		\end{tabular}
	\end{center}
	\caption{VarGFaceNet vs. y2. Performance is recorded within the same epoch. The validation performance of VarGFaceNet is 0.6\% and 0.2\% higher than y2 on AgeDB-30 and CFP-FP respectively. Testing result of VarGFaceNet is 5\% higher than y2.}
	\label{scratch}
\end{table*}
\subsection{Recursive Knowledge Distillation}
\label{sec:kd}
Knowledge distillation with one generation is sometimes difficult to transfer enough knowledge when large discrepancy exists between teacher models and student models. For instance, in our implementation, the FLOPs of teacher model is 24G while that of student model is 1G. And the number of parameters of teacher model is 108M while that of student model is 5M. Moreover, the different architecture and block settings between teacher model and student model increase the complexity of training as well. To improve the discriminative and generalization ability of our student network, we use recursive knowledge distillation, which employs the first generation of student to initialize the second generation of student, as shown in Fig. \ref{kd_f}.

In recursive knowledge distillation, we employ the same teacher model in all generations. That means the angular information of samples which guides the student model is invariable. There are two merits if we use recursive knowledge distillation:

\begin{itemize}
	\item[1] It will be easier to approach guided direction of teacher when a good initialization is applied.
	\item[2] The conflicts between margin of classification loss and guided angular information in the first generation will be relieved in the next generation.
\end{itemize}

The results of our experiments in Section \ref{Experiments} illustrate the performance of recursive knowledge distillation.

\begin{table*}[htb]
	\begin{center}
		\setlength{\tabcolsep}{7mm}
		\begin{tabular}{l|c|c|c|c}
			\hline
			\hline
			Method &  LFW & CFP-FP & AgeDB-30 & deepglint-light (TPR@FPR=1e-8) \\
			\hline
			teacher &  0.99683 & 0.98414 & 0.98083 & 0.86846 \\
			student & 0.99683 & 0.98171 & 0.97550  & 0.84341 \\
			\hline
			teacher & 0.99817 & 0.98729 & 0.98133 & 0.90231 \\
			student & 0.99733 & 0.98200 & 0.98100 & 0.85461 \\
			\hline
			teacher & 0.99833 & 0.99057 & 0.98250 & 0.93315 \\
			student & 0.99783 & 0.98400 & 0.98067 & 0.88334 \\
			\hline
			\hline
		\end{tabular}
	\end{center}
	\caption{Performance of VarGFaceNet with the guide of different teacher models. Performance is recorded within the same epoch.  Results of CFP-FP(validation set) and deepglint-light(TPR@FPR=1e-8) (testing set) show that the higher performance of teacher model leads to the better results of student model.   }
	\label{kd}
\end{table*}

\begin{table*}[h]
	\begin{center}
		\setlength{\tabcolsep}{5mm}
		\begin{tabular}{l|c|c|c|c}
			\hline
			\hline
			Network & LFW & CFP-FP & AgeDB-30 & Flops \\
			\hline
			r100(teacher) & 0.9987 & 0.9917 & 0.9852 & 24G \\
			\hline
			VarGNet(student) & 0.9977 & 0.9810 & 0.9810  & 1029M \\
			\hline
			VarGFaceNet(student) & 0.9985 & 0.9850 & 0.9815  & 1022M  \\
			\hline
			\hline
		\end{tabular}
	\end{center}
	\caption{VarGFaceNet vs. VarGNet. We show the highest performance of every validation dataset. The performance of VarGFaceNet is higher than VarGNet on LFW, AgeDB-30 and CFP-FP.}
	\label{VvV}
\end{table*}

\begin{table*}[h]
	\begin{center}
		\setlength{\tabcolsep}{7mm}
		\begin{tabular}{l|c|c|c|c}
			\hline
			\hline
			Method &  LFW & CFP-FP & AgeDB-30 & deepglint-light (TPR@FPR=1e-8) \\
			\hline
			recursive=1 & 0.99783 & 0.98400 & 0.98067 & 0.88334  \\
			\hline
			recursive=2 & 0.99833 & 0.98271 & 0.98050 & 0.88784 \\
			\hline
			\hline
		\end{tabular}
	\end{center}
	\caption{Performance of recursive knowledge distillation. Performance is recorded within the same epoch.}  Verification results of LFW, AgeDB-30 are increased in the second generation. Performance of testing set deepglint-light(TPR@FPR=1e-8) is increased by 0.4\% the same time.
	\label{rkd}
\end{table*}

\section{Experiments}
\label{Experiments}

In this section, we first introduce the datasets and evaluation metric. Then, to demonstrate the effectiveness of our VarGFaceNet, we compare our network with y2 network(a deeper mobilefacenet\cite{chen2018mobilefacenets,arcface}). After that, the investigation for the effect of different teacher models in knowledge distillation is revealed. Finally, we show the competitive performance of VarGFaceNet using recursive knowledge distillation on LFR2019 Challenge.
\subsection{Datasets and Evaluation Metric}
We employ the dataset(clean from MS1M\cite{guo2016ms}) provided by LFR2019 for training. All face images in this dataset are aligned by five facial landmarks predicted from RetinaFace\cite{deng2019retinaface} then resized to $112\times112$. There are 5.1M images collected from 93K identities. For test set, Trillion-pairs dataset \cite{trillionpairs} is used. It contains two parts: 1) ELFW: Face images of celebrities in the LFW name list. There are 274K images from 5.7K identities; 2) DELFW: Distractors for ELFW. There are 1.58 M face images from Flickr. All test images are preprocessed and  resized to $112\times112$. We refer deepglint-light to trillionpairs testing set in the following. During the training, we utilize face verification datasets (e.g. LFW\cite{huang2008labeled}, CFP-FP\cite{sengupta2016frontal}, AgeDB-30\cite{moschoglou2017agedb}) to validate different settings using 1:1 verification protocol. Moreover, we employ the TPR@FPR=1e-8 as evaluation metric for identification.
\subsection{VarGFaceNet train from scratch}
To validate the efficiency and effectiveness of VarGFaceNet, we first train our network from scratch, and compare the performance with mobilefacenet(y2) \cite{chen2018mobilefacenets,arcface}. We employ arcface loss as the objective function of classification during training. Tabel \ref{scratch} presents the comparison results of VarGFaceNet and y2. It can be observed that under the limitation of 1G FLOPs, VarGFaceNet is able to reach better face recognition performance on validation sets. Compared with y2, our verification results of AgeDB-30 , CFP-FP have increased 0.6\% and 0.2\% respectively, testing result of deepglint-light (TPR@FPR=1e-8) has increased 5\%.  There are two intuitions for the better performance: 1. our network can contain more parameters than y2 when limit FLOPs because of variable group convolution. The biggest number of channels is 256 in y2 while ours is 320 before last conv. 2. Our embedding setting can extract more essential information. y2 expands the number of channels from 256 to 512 then use $7\times7$ depthwise convolution to get the feature tensor before fc layer. We expand the number of channels from 320 to 1024 then use variable group convolution and pointwise convolution which have larger network capacity.

\subsection{VarGFaceNet guided by ResNet}
\label{VgR}
In order to achieve higher performance than train from scratch, bigger networks are applied to perform knowledge distillation using angular distillation loss. Moreover, we conduct experiments to investigate the effect of different teacher models on VarGFaceNet. We employ ResNet 100 \cite{resnet} with SE as our teacher model. The teacher model has 24G FLOPs and 108M parameters. The results are illustrated in Tabel \ref{kd}. It can be observed that 1. even though the architectures of teacher and student are quite different, VarGFaceNet still approaches the performance of ResNet; 2.  the performance of VarGFaceNet is highly correlated with teacher model. The higher performance teacher model has, the better interpretation ability VarGFaceNet will learn.

To validate the efficiency of our settings, we conduct comparison experiments between our network and \textit{VarGNet}. Using the same teacher network, we change the head setting of \textit{VarGNet} to our head setting for fair comparison and use the same loss function as above. In Tabel \ref{VvV}, the plain \textit{VarGNet} has lower accuracy in LFW, CFP-FP, AgeDB-30. There is only an average pooling between last conv and fc layer in \textit{VarGNet}. The results illustrate that our embedding setting is more suitable for face recognition task since it can extract more essential information.

\subsection{Recursive Knowledge Distillation}
As we discuss in Section \ref{sec:kd}, when there is a large discrepancy between teacher model and student, knowledge distillation for one generation may not enough for knowledge transfer. To validate it, we use ResNet 100 model as our teacher model, and conduct recursive knowledge distillation on VarGFaceNet. A performance improvement shown in Table \ref{rkd} when we train the model in next generation. The varification result of LFW and CFP-FP is increased by 0.1\% while testing result of deepglint-light(TPR@FPR=1e-8) is 0.4\% higher than pervious generation. Furthermore, we believe that it will lead to better performance if we continue to conduct training in more generations.

\section{Conclusion}
In this paper, we propose an efficient lightweight network called VarGFaceNet for large scale face recognition. Benefit from variable group convolution, VarGFaceNet is capable of finding a better trade-off between efficiency and performance. The head setting and embedding setting specific to face recogniton help preserve information while reduce parametes. Moreover, to improve the interpretation ability of lightweight network, we employ an equivalence of angular distillation loss as our objective function and present a recursive knowledge distillation strategy. The state-of-the-art performance on LFR challenge demonstrates the superiority of our method.

\textbf{Acknowledgments} We would like to thank Xin Wang, Helong Zhou, Zhichao Li, Xiao Jiang, Yuxiang Tuo for their helpful discussion, especially Helong for his advice and discussion on recursive knowledge distillation.

{\small
\bibliographystyle{ieee}
\bibliography{egbib}
}

\end{document}